\documentclass[11pt]{article}
\usepackage{coling2020}
\usepackage{times}
\usepackage{url}
\usepackage{latexsym}
\usepackage{booktabs}
\usepackage{graphicx}  % DO NOT CHANGE THIS

\usepackage{amsmath} % AUTHOR ADDED PACKAGE
\usepackage{booktabs} % AUTHOR ADDED PACKAGE

\colingfinalcopy % Uncomment this line for the final submission

\title{Improving Human-Labeled Data through Dynamic Automatic Conflict Resolution}
\author{David Q. Sun\textsuperscript{*}, Hadas Kotek\textsuperscript{*}, Christopher Klein,  Mayank Gupta, William Li, Jason D. Williams \\ % All authors must be in the same font size and format. Use \Large and \textbf to achieve this result when breaking a line
\\ Apple \\ One Apple Park Way\\
Cupertino, California 95014\\
dqs@apple.com % email address must be in roman text type, not monospace or sans serif
}

\usepackage[symbol]{footmisc}
\renewcommand{\thefootnote}{\fnsymbol{footnote}}

\begin{document}
\maketitle

\footnotetext[1]{These authors contributed equally to this work.}

\renewcommand{\thefootnote}{\arabic{footnote}}
\setcounter{footnote}{0}

\begin{abstract}

This paper develops and implements a scalable methodology for (a) estimating the noisiness of labels produced by a typical crowdsourcing semantic annotation task, and (b) reducing the resulting error of the labeling process by as much as 20-30\% in comparison to other common labeling strategies. Importantly, this new approach to the labeling process, which we name Dynamic Automatic Conflict Resolution (DACR), does not require a ground truth dataset and is instead based on inter-project annotation inconsistencies. This makes DACR not only more accurate but also available to a broad range of labeling tasks. In what follows we present results from a text classification task performed at scale for a commercial personal assistant, and evaluate the inherent ambiguity uncovered by this annotation strategy as compared to other common labeling strategies.

\end{abstract}

\section{Introduction}

Ground truth datasets for supervised learning are commonly collected through crowdsourcing annotation projects, either on public platforms (e.g. Yandex.Toloka, Amazon Mechanical Turk, Figure Eight) or on in-house counterparts. In both scenarios, the annotation project is conducted by first providing workers with annotation guidelines and then assigning each one with a set of annotation requests. Each unique annotation request is usually assigned to more than one worker, to increase the reliability of the results. Subsequently, some post-processing technique is used to create a consensus label that is recorded as the \textit{ground truth}.

%There are two major deficiencies to this approach to determining the ground truth, however: first, this approach overlooks latent inaccuracies \fixme{in .. what}; second, it ignores ambiguities inherent to the process itself. 
However, since human annotations are intrinsically noisy and subject to the influence of external factors, such a process is unlikely to produce accurate ground truth labels for 100\% of the requests. Moreover, the practice of using a singular consensus label for multi-annotated requests mandates a \textit{one and only} correct interpretation for every request in any given task, effectively denying the existence of potential ambiguities. This implicit \textit{denial of ambiguity}, while acceptable for certain tasks such as arithmetics, can be problematic in tasks such as natural language understanding \cite{dumitrache2018capturing}. Perhaps because of these difficulties, estimating the latent accuracy of any annotation process has been a relatively under-examined subject.

In this paper, we first introduce a scalable methodology to estimate the latent inaccuracy of any crowdsourcing annotation process. The estimation method does not require any preexisting ground truth dataset, and is therefore immune to the \textit{Gold Question Inference} type of attacks described by Checco et al.~\shortcite{checco2018all}. We construct an agent-based model to examine the impact of different annotation strategies on overall annotation accuracy. Inspired by the findings of our simulation experiments, we propose and implement a novel process of \textit{Dynamic Automatic Conflict Resolution (DACR)}, designed to improve the accuracy of the annotation output with limited incremental workload while collecting sufficient annotation input for requests with ambiguity. We present annotation results from a DACR project used in the context of a text classification task, implemented at scale for a commercial personal assistant.

The primary contributions of this paper are:

\begin{enumerate}
	\item \textit{latent inaccuracy estimation methodology}: we propose a simple, robust, scalable methodology and theoretical framework to study the latent inaccuracies of any crowdsourcing annotation process;
	\item \textit{comparative study of different annotation strategies}: we conduct an in-depth comparison of the performance of commonly used annotation strategies via agent-based simulation experiments;
	\item \textit{DACR, an ambiguity-preserving and accuracy-boosting annotation strategy}: we introduce the Dynamic Automatic Conflict Resolution strategy to improve the accuracy of the annotation process;
	\item \textit{empirical analysis on a text classification task}: we present statistical findings from a DACR annotation project performed at scale on a text classification task.
\end{enumerate}

\section{Related Work}

This work relates to recent developments in three areas of research: (1) efficiency and accuracy of human annotation (in particular, inter-annotator agreement), (2) quality assurance in crowdsourcing, and (3) inherent ambiguity in natural language annotation tasks. We survey these in turn below.

\subsection{Evaluating Human Annotation}

Annotated natural language data is invaluable to machine learning techniques and algorithms. Two major questions arise as to how to optimize data collection task and how to measure the accuracy and reliability of the resulting data.

Many authors emphasize the importance of careful task design, guideline creation, and pre-testing as seen in~\cite{compton2012developing,compton2012developing,pustejovsky2012natural}. Such tests assist in improving \textit{inter-annotator reliability} --- the extent to which the results of an annotation task represent replicable unbiased judgments. Several reliability estimates correct for this, see especially Krippendorff's $\alpha$ \cite{krippendorff2004reliability}.

Once data is collected, it is standard practice to take \textit{Inter-Annotator Agreement (IAA)} to provide an estimate of the correctness or validity of human annotations (\cite{smyth1995learning}). IAA is distinct from the reliability metric: the former cannot account for biases that might be shared by a group and hence lead to high agreement across annotators. There is no consensus on how to measure IAA, but one common metric is the Kappa ($\kappa$) statistic (or some variant), which measures pairwise agreement among a set of annotators making category judgments and correcting for expected chance agreement: 

\begin{center}
	$\kappa = \frac{Pr(a) - Pr(e)}{1 - Pr(e)}$,
\end{center}

\noindent where $Pr(a)$ is the proportion of times that the annotators agree and $Pr(e)$ is the proportion of times that we would expect them to agree by chance.

This measure has been shown to be effective, but it also has certain limitations~\cite{eugenio2004kappa,pustejovsky2012natural,artstein2017inter}. See also~\cite{craggs2005evaluating,passonneau2014benefits,kara2018actively} for some less commonly used measures.

In addition to the ongoing debate about the accuracy and correctness of the $\alpha$ and $\kappa$ statistics as proxies for inter-annotator reliability and agreement respectively, the metrics provide only a partial picture of human-annotated data. The statistic is unable to account for the level of expertise of the annotators, for example. In a meta study, \cite{bayerl2011determines} identify several factors that may influence IAA: annotation domain, number of categories in a coding scheme, number of annotators in a project, whether annotators received training, the intensity of annotator training, the annotation purpose, and the method used for calculating IAA. One of their key conclusions is that complex annotation tasks should be simplified to use fewer grading categories. \cite{bayerl2011determines} additionally suggest using expert annotators who receive specialized training and collecting a large number of annotations for critical tasks. 

The use of expert annotators versus naive ones is a contested issue. While some authors suggest using experts or adjudicators \cite{kowtko1993conversational}, Snow et al.~\shortcite{snow2008cheap} argue that many labeling tasks can be effectively designed and implemented using non-expert annotators recruited via crowd-sourcing tools.

\subsection{Quality Assurance in Crowdsourcing}

Many solutions exist to resolve the low-quality contributions that come either from deliberate scamming or poor performance \cite{ipeirotis2010quality,dow2012shepherding,ipeirotis2014repeated,gadiraju2015understanding}. The most commonly adopted technique for quality assurance relies on a set of questions with known correct answers, often referred to as \textit{gold standard test questions}, which are randomly injected into the workers' sessions in order to measure their performance \cite{checco2018all,sun2012majority,el2014skill}.
In addition, since the original Dawid-Sekene \cite{dawid1979maximum} crowdsourcing model based on maximum likelihood estimation, several improvements have been proposed:  the Fast Dawid-Skene model for improved speed of convergence \cite{sinha2018fast}, the effort to extend to multiclass labeling from binary labeling \cite{li2014error}, and the Minmax Entropy method \cite{zhou2012learning} for accuracy improvements. There are also other comparable models \cite{raykar2010learning,welinder2010multidimensional,karger2011iterative}, but none could outperform Dawid-Skene \cite{liu2012variational}.

However, such techniques share several clear limitations:
\begin{enumerate}
	\item \textit{Scalability Challenge}: the curation of gold standard test questions, or the initialization for ground truth in maximum likelihood estimation methods, often demands significant effort and resources (as many as 30 identical batch of graders per question assignment), and therefore such question sets are usually small in size \cite{oleson2011programmatic};
	\item \textit{Measurement by Proxy}: the gold standard test questions only measure the annotator performances on the given question set, and the results do not necessarily reflect their performances on the actual tasks; further, the question set may differ significantly in terms of style or difficulty from the requests in the actual annotation task, leading to greater deviation from the intended measurement;
	\item \textit{Vulnerability to Exploitation}: the inherent limit on the size of the gold standard test question set makes the method prone to a coordinated attack from even a small group of annotation workers, once they can collude and identify the gold questions; Checco et al.~\shortcite{checco2018all} have proposed an attack mechanism targeting the gold question-based quality assurance method.
\end{enumerate}

\subsection{Inter-Annotator Agreement and Ambiguity}
This work is in part a continued discussion on inter-annotator (dis)agreement and its implications. As suggested by many~\cite{dumitrache2018empirical,cheatham2013string,plank2014learning,hollenstein2016inconsistency}, the lack of consensus among a group of annotators on a specific task can be attributed to one or more of the following causes: inherent uncertainty in the relevant domain knowledge, true natural language ambiguity, and annotation error. Here we do not attempt to provide a comprehensive interpretation of inter-annotator disagreement. Instead, we propose an annotation strategy that  efficiently preserves material inter-annotator disagreement while reducing the impact of annotation errors on consensus extraction. We then share findings from a text classification task which uses this strategy and examine some specific cases of inter-annotator disagreement. We show that the DACR strategy can assist  in identifying material disagreements --- ones which stem from issues with domain knowledge or the ontology used in the classification task --- while reducing disagreements related to simple human error.

\section{Latent Inaccuracy Estimation Method (LIEM)}

This section presents a mathematical formula for measuring the inaccuracy of a human annotation task.

\subsection{LIEM: a brief description}
 
The LIEM treats a given annotation process as a black box system with the following assumptions:

 \begin{itemize}
 	\item{\textit{Singular ground truth label}: each request has a \textit{singular} ground truth label;}
 	\item{\textit{Absence of systematic confusion}: the annotation process yields \textit{incorrect} labels from a discrete uniform distribution and doesn't contain systematic confusions;}
 	\item{\textit{Memoryless behavioral stability}: the annotation process is not affected by \textit{a priori} events, and exhibits stable behavior over time}.
 \end{itemize}

Given these assumptions, we may assign labels to a set of requests using the same predetermined annotation process in two duplicate annotation projects, and compute the proportion of requests that received identical labels, $\hat{Y}$. The latent accuracy may then be expressed as:

\begin{center}
	$\mu_X \approx \sqrt{\hat{Y}}  $,
\end{center}

\noindent with decreasing margin of error for larger values of $n$.
 
\subsection{LIEM: a detailed derivation}

In the remainder of this section we explicate this result in detail. To start, consider a project with an unspecified annotation task for a set of requests $R$. Let us assume each request $r\in R$ has a singular \textit{ground truth} label $g_r\in\Omega$, where $\Omega$ is a finite set of all permissible labels in the annotation task.\footnote{Although we acknowledge potential ambiguity and existence of multiple correct labels, this stipulation allows us to study the impact of annotation inaccuracy for the prevalent choice of singular ground truth label in semantic annotation tasks.}

\subsection{Expected performance of a black box system}

Each request $r$ undergoes a predetermined annotation process, which returns a label $l_r\in\Omega$. In the crowdsourcing context, the returned label $l_r$ is extracted from the answer(s) provided by one or more \textit{randomly selected} annotation workers. Therefore, if we treat the predetermined annotation process as a black box system and focus on its input-output relationship, the system is expected to behave inconsistently in terms of its performance (that is, its ability to yield the correct answer such that $l_r=g_r$) due to (1) varying difficulty levels of each request and (2) non-uniformity of performance both amongst annotation workers and for any individual worker over different requests.

Let $X$ be a continuous random variable over $[0,1]$ associated with the performance of the predetermined annotation process. The performance of the system for each request $r$ is defined as the likelihood of yielding the correct label, i.e. $P(l_r=g_r)$. Accordingly, we consider the performance of the system, $p_{r,t}\in[0,1]$, for each request $r$ at time $t$ to be an independent random draw from $X$. We denote the \textit{expected performance} of the system as $E[X] = \mu_X$, the value we intend to estimate.

\subsection{Central limit theorem and Bhatia–Davis inequality}

Next, we briefly discuss two important concepts of probability: the \textit{Central Limit Theorem (CLT)} and \textit{Bhatia-Davis inequality}. More complete introductions can be found in \cite{barron1986entropy,bhatia1993more}. Without loss of generality, we may treat the performances $\{X_1, X_2,... X_n\}$ of a black box system over a set of $n$ annotation requests (where $n = |R|$) as a \textbf{random sample} of size $n$ from the given distribution of $X$, with expectation $E[X_i]=\mu_X$ and variance $Var[X_i]=Var[X]=\sigma_X^2$. Such a set of random variables are \textbf{independent, identically distributed}. Conveniently, then, the sample mean:

\begin{center}
	 $\overline{X}=\frac{1}{n} \sum_{i=1}^nX_i = \frac{1}{n}\sum_{i=1}^n1\cdot X_i$,
\end{center}

\noindent is the aggregated distribution of the proportion of correctly labeled requests in $R$. Based on CLT, we conclude that the mean of a sufficiently large random sample from an \textit{arbitrary} distribution to have approximately normal distribution such that $\overline{X} \sim \mathcal{N}(\mu_X, \sigma_X^2/n)$.\footnote{The larger the sample size, the better the CLT approximation becomes.} With the help of the Bhatia-Davis inequality, we may write the upper bound on the variance of any \textit{bounded} distribution on the real line as:

\begin{center}
	 $\sigma^2\leq (M-\mu)(\mu - m)$,
\end{center}

\noindent where $M$, $m$ are the maximum and minimum of the distribution. For $X$ with $M=1, m=0$ we have:

\begin{center}
	 $\sigma_X^2\leq (1-\mu_X)\mu_X$.
\end{center}

The method to estimate the accuracy of the predetermined annotation process is trivial in the case when $R$ is a gold standard question set so that \textit{the proportion of correctly labeled requests} is directly observable. Since the premise of this work is to estimate accuracy without relying on gold questions, we continue on to next section.

\subsection{Consistency as a reliable proxy}

Turning back to our labeling task, in order to find a reliable proxy variable, we examine the annotation process for a single request $r$ in detail. Given the finite set of all permissible labels in the annotation task, $\Omega$,  for request $r$ with ground truth label $g_r\in \Omega$, the likelihood of receiving the correct label such that $l_r = g_r$ is previously defined as the \textit{performance} of the annotation process on $r$, i.e. $X_r \equiv P(l_r=g_r)$.

Accordingly, the probability of not receiving a correct label is $P(l_r \in \Omega \setminus \{g_r\}) = 1 - P(l_r = g_r)$. Now, let us assume that when the annotation process (treated as a \textit{black box system}) does not yield the correct label, it simply returns one of the remaining labels in the sample space, i.e. $\Omega \setminus \{g_r\}$, with equal probability\footnote{\cite{giancola2018permutation} suggest that \textit{mislabeling} is not necessarily follow a discrete uniform distribution, and systematic confusions (aptly named "class permutations") are observed among the annotation workers on several tasks including text clustering, rare class simulation, and dense image segmentation. However, their findings from the text clustering task also suggest that such systematic confusions are relatively less prevalent in the given task. We assume a discrete uniform distribution as a necessary approximation to avoid \textit{intractability}.}.

Consider a request $r$ undergoing the said annotation process \textit{twice}, with the system's performances at $X_r = p, X_r' = q$, yielding labels $l_r,l_r'$ respectively. With the help of Bayes' theorem, we may write the probability of receiving two identical labels as: \\
\newline
 $P(l_r=l_r') = $ \\ 
 $P(l_r=l_r' | l_r = g_r, l_r' = g_r)\cdot P(l_r=g_r) \cdot P(l_r'=g_r) + P(l_r=l_r' | l_r \neq g_r, l_r' \neq g_r) \cdot P(l_r \neq g_r) \cdot P(l_r' \neq g_r)$, \\
\newline
\noindent from which we calculate: 

\begin{center}
	 $P(l_r=l_r') = p\cdot q + \frac{1}{m-1} \cdot (1-p)(1-q) $,
\end{center}

\noindent where $m = | \Omega |$. 
 
 Further, with reasonably good faith in the system performances such that $pq \gg (1-p)(1-q)$ and a sufficiently large sample space (e.g. $m>10$), we may allow for further approximation given a significantly smaller second term\footnote{Essentially, the chances of receiving identical labels from two independent annotation sessions on the same request are negligible when both labels are incorrect.} as $\frac{1}{m-1} \cdot (1-p)(1-q) \rightarrow 0$ and $p\cdot q \gg \frac{1}{m-1} \cdot (1-p)(1-q)$.  This approximation enables us to simplify the expression for  $P(l_r=l_r')$ to:
 
\begin{center}
 	$P(l_r=l_r') = p\cdot q$,
\end{center}

\noindent based on which we may express the probability distribution of receiving identical labels for a request $r$ in two independent tries from a predetermined annotation process as a \textit{product distribution},

\begin{center}
	$Y_R = X_R X_R'$,
\end{center}

\noindent with its mean $E[Y_R] = E[X_R]\cdot E[X_R'] = \mu_X^2$ and variance $Var[Y_R] = (2\mu_X^2 + \sigma_X^2)\sigma_X^2$.\footnote{The mathematical details to derive the variance of the product of two independent random variables can be found in \cite{goodman1960exact}} 

\subsection{Duplicate annotation: measuring consistency}

To measure consistency at scale, we set up two annotation projects with an identical set of requests $R$. This yields two sets of labels, $L = \{l_1, l_2,...,l_n\}, L' = \{l_1',l_2',...,l_n'\}$. We may then treat the consistency observations $\{Y_1, Y_2,...,Y_n\}$ as a random sample of size $n$ from a distribution $Y$, with expectation $E[Y] = \mu_X^2$ and variance $Var[Y] = (2\mu_X^2 + \sigma_X^2)\sigma_X^2$. Assuming that they are independent, identically distributed, according to Central Limit Theorem the distribution of sample mean converges to a normal distribution when the sample size is large enough:  

\begin{center}
	$\overline{Y} \overset{\mathrel{D}}{\longrightarrow} \mathcal{N}(\mu_X^2, (2\mu_X^2 + \sigma_X^2)\sigma_X^2/n)$,
\end{center}

\noindent where by Bhatia–Davis inequality, we know $\sigma_X^2\leq (1-\mu_X)\mu_X$, so that $\max\{(1-\mu_X)\mu_X\} = \frac{1}{4}$ for $\mu_X \in [0,1]$. Accordingly, we may establish a (loose) upper bound for the variance of $\overline{Y}$ as:

\begin{center}
	$Var[\overline{Y}] = (2\mu_X^2 + \sigma_X^2)\sigma_X^2/n \leq \frac{33}{64n}$,
\end{center}

\noindent from which we find that $Var[\overline{Y}] \rightarrow 0$ as $n \rightarrow \infty$, i.e. by \textit{law of large numbers}, the observed sample mean becomes closer to $E[Y]$ as the sample size increases.

Meanwhile, the established upper bound helps us to properly capture the margin of error in estimating $E[Y]$ for a not-so-large size of $n$. For example, with $n=100$ we have $Var[\overline{Y}] \leq 5.16\cdot 10^{-3}$; to maintain a confidence level of a three-sigma effect (99.73\% confidence) we would estimate $\mu_X^2$ to be $\overline{Y} \pm 0.015$.

\subsection{LIEM: a summary}
 
In sum, the Latent Inaccuracy Estimation Method treats a given annotation process as a \textit{black box system} and assumes that (a) each request has a singular ground truth label; (b) the annotation process yields incorrect labels from a discrete uniform distribution and doesn't contain systematic confusions; and (c) the annotation process is not affected by \textit{a priori} events, and exhibits stable behavior over time.

Therefore, if we assign labels to a set of requests using the same annotation process in two duplicate projects, we may compute the proportion of requests that receive identical labels, $\hat{Y}$. The latent accuracy of the annotation process may then be expressed as: $\mu_X \approx \sqrt{\hat{Y}}$, with decreasing margin of error for larger values of $n$.

\section{A Look into Different Annotation Strategies}

With this result in hand, we examine three commonly used annotation strategies and evaluate their respective performances through agent based simulations. We compare these results to our novel methodology, Dynamic Automatic Conflict Resolution (DACR), which we introduce immediately below. We establish that DACR outperforms other common approaches to human annotation. In a subsequent section we will show that DACR indeed outperforms these other approaches in an actual annotation task.

\subsection{Annotation Strategies}

We compare the following three commonly used annotation strategies:

\begin{enumerate}
	\item \textit{One-grader}: each request is assigned to a randomly selected grader and the answer label is recorded;
	\item \textit{Double Graded, Conflict Resolved (DG, CR)}: each request is first assigned to \textit{two} randomly selected graders; if the two agree, the consensus answer label is recorded; otherwise (in the case of a conflict), the request is escalated to an \textit{expert annotator} with higher expected performance, and the expert's answer label is recorded as the final answer;
	\item \textit{N-Graded}: each request is assigned to \textit{N} randomly selected graders; the answer label with a simple majority (greater than $\frac{1}{2}$) is recorded as the final answer. If no answer label with a simple majority exists, the request is assigned an \textit{in-conflict} label.
\end{enumerate}

Against these approaches, we propose a novel approach: \textit{Dynamic Automatic Conflict Resolution (DACR)} with maximum $N$ grades: each request is first assigned to \textit{two} randomly selected annotation workers; if the two agree, the consensus answer label is recorded; otherwise (in the case of a conflict), the request is iteratively assigned to one additional worker until (a) a simple majority answer label (from all historical answers) is produced, or (b) the maximum $N$ grades is reached and the annotation process ends (regardless of the presence of a majority label). A separate process can then apply to determine how to resolve \textit{in-conflict} requests.

\begin{table}[!h]
\centering
\begin{tabular}{@{}lll@{}}
\toprule
Parameter                 & Initialization          & Count \\ \midrule
Annotation Workers        & $\mathrm{unif}(0.8, 1)$ & 100  \\
Expert Annotators         & $\mathrm{unif}(0.9, 1)$ & 20  \\
Request Difficulty Levels & $\min(1,\mathcal{N}(0.9, 0.1))$ & Varied  \\ \bottomrule
\end{tabular}
\caption{{\em Initialization configurations}}\vspace{-1em}
\label{tab:initial_config}
\end{table}

\subsection{Simulation Configurations}

Our simulation experiments vary the following parameters of the crowdsourcing annotation task: (1) the quality of the annotation workers, (2) the quality of expert annotators (if any), (3) the difficulty levels of the requests in the annotation task, and (4) the annotation strategy in use.

As detailed in Table~\ref{tab:initial_config}, the parameters are initialized{\footnote{We also simulated with varying parameter values and the subsequent findings  are established consistently.}} with reasonable estimates from empirical observations. Accordingly, the annotation act is abstracted as follows: when an annotator with capability $c$ annotates a request with difficulty level $d$, the correctness of the answer label is generated as the outcome of a single \textit{Bernoulli trial}, with $P(correct)=d\cdot c$, and $P(incorrect)=1- (d\cdot c)$.

\subsection{Experiment design and findings}

Our simulation experiments investigate (1) the \textit{performance} and \textit{efficiency} of the competing annotation strategies outlined above, and (2) the validity of the proposed LIEM approach in estimating the expected performance of a given annotation strategy.

\begin{figure}[!h]
\centering
\includegraphics[width = 0.5 \textwidth]{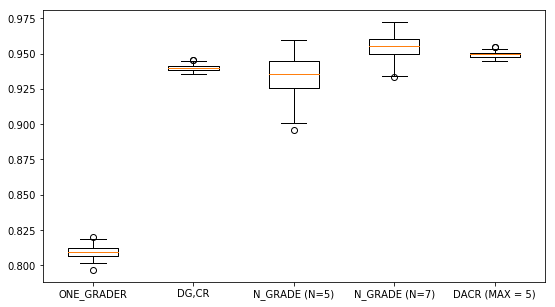}
\caption{The distribution of annotation accuracy from 100 simulated annotation projects on an identical batch of 10,000 requests, produced by different annotation strategies.}
\label{fig:10K_acc_simulations} \vspace*{-0.25em}
\end{figure}

We ran 100 simulated annotation projects for each annotation strategy; each project consisted of the same batch of randomly initialized 10,000 requests. We plot the distribution of the annotation accuracy over the projects for each strategy in Figure~\ref{fig:10K_acc_simulations}. 

We  conclude in Table~\ref{tab:annotation_performances} that the DACR strategy is consistently ranked top 2 compared to its competitors along all three metrics listed. We furthermore find that the DACR strategy requires a \textit{significantly lower} number of grades to achieve comparable performance to that produced by a 7-graded strategy (63.6\% lower than this top competitor), and achieves improved performance with a small incremental number of grades compared to the $DG, CR$ strategy (9.0\% higher accuracy compared to this top competitor).

\begin{table}[!h]
\centering
\begin{tabular}{@{}llll@{}}
\toprule
Strategy                 & $\overline X$          & $\sigma^2$ 		& Avg. Grades\\ 
\midrule
One-grader         & 80.9\% & $1.47e^{-5}$  & 10,000		\\
DG, CR         		& 94.0\% & $\mathbf{4.65e^{-6}}$  & \textbf{23,382}		\\
N-grader (N=5)         & 93.4\% & $1.92e^{-4}$  & 50,000	\\
N-grader (N=7)         & \textbf{95.4\%} & $7.12e^{-5}$  & 70,000	\\
\textbf{DACR} (Max = 5) & \textbf{94.9\%} & $\mathbf{3.74e^{-6}}$  	& \textbf{25,495}		\\
\bottomrule
\end{tabular}
\caption{{\em Sample mean, variance and average total grades, from 100 simulated annotation projects on an identical set of 10,000 requests, produced by different annotation strategies}}\vspace*{-0.25em}
\label{tab:annotation_performances}
\end{table}

\begin{figure}[!h]
\centering
\includegraphics[width = 0.4 \textwidth]{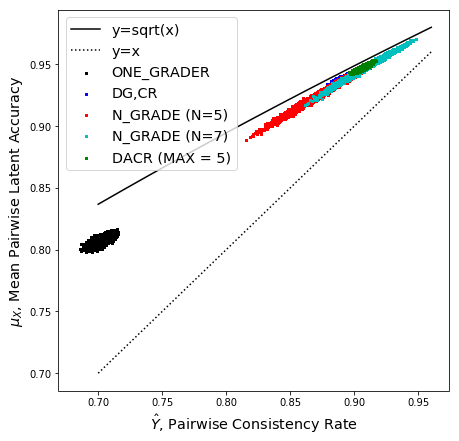}
\caption{Inter-project consistency rates, mean pairwise accuracy for different annotation strategies; proximity to the solid line $y=\sqrt{x}$ suggests LIEM is a reasonable estimation method, where $\mu_X \approx \sqrt{\hat{Y}}$.}
\label{fig:ca_pairs} \vspace{-0.5em}
\end{figure}

To validate the LIEM approach, we exhaustively compute the inter-project consistency rates and the corresponding pair's mean accuracy, for each annotation strategy. The data points are plotted in Figure~\ref{fig:ca_pairs}, where we conclude with simple curve fitting that LIEM's proposed estimation method, $\mu_X \approx \sqrt{\hat{Y}}$, is a reasonably accurate approach to find the latent accuracy for a variety of annotation strategies.

\section{Exploring Ambiguity with DACR}

As a part of a greater effort to build an intelligent virtual assistant, we have long been working on a text classification task. The text classifier predicts user intent for a given utterance according to a predetermined taxonomy, as in Table~\ref{tab:example_utterances}. In our task, over 60 possible intent labels are available and we stipulate that only one correct intent label exists for any given utterance. In the remainder of this paper, we introduce the results from a DACR project implemented at scale for a commercially used assistant.

\begin{table}[h!]
\centering
\begin{tabular}{@{}l|l@{}}
\toprule
User Utterance                 & Intent Class \\ \midrule
Davos local time      & Time  \\
Status for AA 166         & Flight  \\
Compute 37 times 30 & Math   \\ 
Kesha album Rainbow & Music   \\ 
\bottomrule
\end{tabular}
\caption{{\em Example utterances with their intent labels}}
\label{tab:example_utterances} \vspace{-0.75em}
\end{table}

 \subsection{Results from DACR} 

Having examined DACR's (superior) performance via simulation experiments, we deployed it to our internal crowdsourcing platform for annotation on a text classification task. The project consisted of 60,000 randomly sampled utterances and used the DACR annotation strategy, with the minimum grade count set to 3 and the maximum set to 9. At the completion of the project, we recorded an average of 3.74 grades per utterance. Additionally, 5.00\% of the utterances did not receive a consensus label even when the maximum grade count of 9 had been reached (we will refer to such cases as ``\textit{in-conflict}'').\footnote{In simulation, we record 2.55 average grades per utterance with $\mathbf{Min=2}$ and $\mathbf{Max=5}$ , and an \textit{in-conflict} rate of 5.1\%.}
 
\begin{figure}[!h]
\centering
\includegraphics[width = 0.4 \textwidth]{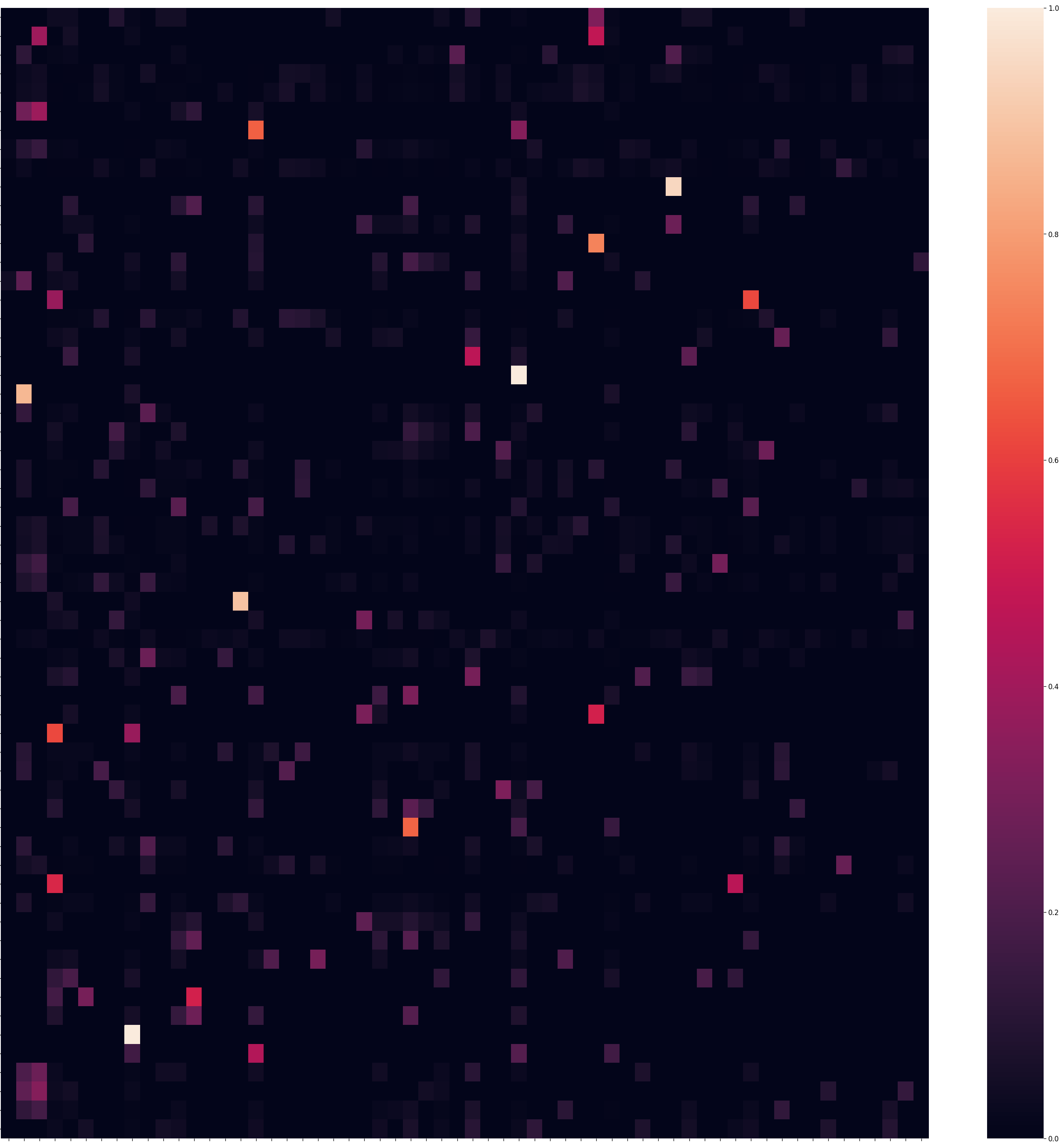}
\caption{Normalized confusion matrix of inter-annotator disagreements over \textit{in-conflict} requests from a text classification task, the exact intent labels are redacted.}
\label{fig:confusion_matrix}\vspace{-0.75em}
\end{figure}

To determine whether such \textit{in-conflict} requests are a result of inherent ambiguity or poor performance of annotation workers, we construct a confusion matrix based on the inter-annotator disagreements over the set of requests with no consensus label. As we observe from Figure~\ref{fig:confusion_matrix}, although we omit the label names to preserve privacy --- there is a clear pattern where a small number of specific intent labels are significantly more likely to be confused with a small set of select other labels.\footnote{Notice the existence of highlighted blocks in Figure~\ref{fig:confusion_matrix}. Such highlighted blocks correspond to unusually high levels of inter-category confusion, and are to be rarely observed in the confusion matrix if no systematic bias exists. The confusion matrix is normalized by adjusting for background frequency of each category.}

 Accordingly, the observed systematic bias indicates the presence of one or several of following factors: (1) ambiguity or vagueness in the annotation guidelines, (2) inherent ambiguity in the ontology design, or (3) inherent ambiguity in a few select types of utterances.

An examination of the \textit{in-conflict} cases has led us to conclude that in this study, (2) and (3) were sizable contributing factors. As a result, we have been able to undertake a series of directed efforts aimed at educating the annotators on the one hand, and better-defining the ontology on the other, to reduce conflicts caused by reason (2). On the other hand, we do not expect that any effort on our part can resolve conflicts casued by reason (3). Instead, the prevalence of truly ambiguous utterances in our study provides yet more evidence that the common insistence on a \textit{single} ground truth label for each request may simply be misguided. In other words, such an assumption makes it inherently impossible to correctly represent 100\% of natural language utterances. 
 
 Acknowledging the existence of ambiguity has two profound implications for nearly all annotation tasks that conforms to the premises of this study:

 \begin{itemize}
 	\item \textit{Information loss}: as \cite{dumitrache2018capturing} have previously suggested, the rejection of ambiguity results in a loss of information that would impact a wide range of supervised learning applications and prevents the development of appropriate model architectures;
 	\item \textit{Measurement error}: we find it equally concerning that if we are to generate a test set with annotation strategies that ignores ambiguity (such as one-grader, or \textit{DG,CR}), we would have deliberately introduced noise into the test set by introducing ambiguous examples with an uncertain partial label.  This omission would lead to increased measurement error on model performance.\footnote{Consider a test set with 5\% ambiguous examples for which no (single) ground truth exists. Then even the \textit{oracle} would be measured to have a 5\% error rate on this test set.}
 \end{itemize}

%\fixmes{add some discussion about how this helps us identify problematic domains; the confusion matrix isn't very informative but we can say something about a subsequent process of identifying top confusions and trying to streamline them. E.g. I've worked on helping improve nonsense. This was also informative in my process of establishing a new weekly assessment project, where we intentionally pair the ``confused'' domains with each other to help the annotators get better at distinguishing them. We're working on eliminating encyclopedia and moving all knowledge to answerFacts. we've deprecated hotelEvent; mealEvent is next. Kyle and I are exploring politics after that. }
% 
\section{Conclusion}

In this paper, we have introduced a method to estimate the latent inaccuracy of any crowdsourcing annotation process and shown its validity both via formal analysis and simulation experiments. We also investigated the performances and efficiencies of different annotation strategies with an agent-based model. Based on our findings, we further propose a novel annotation strategy that dynamically allocates additional annotation resources to difficult (or ambiguous) requests; it is found to be able to preserve multiple conflicting grades for ambiguous requests and outperform comparable annotation strategies while requiring significantly less annotation resources. 

We share insights from a DACR-based annotation project on a classification task and discover patterns among the \textit{in-conflict} examples that corroborate the existence of ambiguity in the given task. Our goal is to allow the practitioners in the field of machine learning to easily estimate the latent accuracy of any crowdsourcing annotation process. We further hope many will find DACR to be a better alternative to their existing annotation strategies and start thinking about how we should deal with inherent ambiguity.

\clearpage

\noindent 
\bibliographystyle{coling}
\bibliography{Bibliography-File}

\end{document}